%% file: paper.tex
\pdfoutput=1
\documentclass{llncs}

\usepackage[utf8]{inputenc}
\usepackage[english]{babel}

\usepackage{amsmath}
\usepackage{amsthm}

\usepackage[T1]{fontenc}
\usepackage[scaled=0.8]{beramono}
\usepackage{latexsym,amsmath,amsfonts,amssymb,braket}
\usepackage{array}
\usepackage{listings}
\usepackage{ifmtarg}
\usepackage{calc}

\usepackage{amsfonts}
\usepackage{amssymb}
\usepackage{graphicx}
\usepackage{caption}
\usepackage{subcaption}
\graphicspath{{../images/}}
\usepackage{lmodern}
\usepackage{url}
\usepackage[noend]{algpseudocode}
\usepackage[table]{xcolor}
\usepackage[labelfont=bf]{caption}
\usepackage{listings}
\usepackage{courier}
\usepackage{enumitem}
\usepackage{emptypage}

\usepackage{siunitx}
\usepackage[table]{xcolor}
\usepackage{booktabs}
\usepackage{multirow}
\usepackage[ruled,vlined,linesnumbered]{algorithm2e}
\SetAlgoProcName{Class}{anautorefname}
\DontPrintSemicolon
\definecolor{vlgray}{gray}{0.92}

\newcommand\Zero{0^{64}}

\newcommand{\f}[1]{\mathtt{#1}} 
\newcommand{\m}[1]{\ProcNameSty{#1}} 

\definecolor{Gray}{gray}{0.9}
\definecolor{mygray}{rgb}{0.1,0.1,0.1}

\lstdefinelanguage{scala}{
  morekeywords={abstract,case,catch,class,def,%
    do,else,extends,false,final,finally,%
    for,if,implicit,import,match,mixin,%
    new,null,object,override,package,%
    private,protected,requires,return,sealed,%
    super,this,throw,trait,true,try,%
    type,val,var,while,with,yield, Array},
  otherkeywords={=>,<-,<\%,<:,>:,\#,@},
  sensitive=true,
  morecomment=[l]{//},
  morecomment=[n]{/*}{*/},
  morestring=[b]",
  morestring=[b]',
  morestring=[b]"""
}

\title{Compact-Table: Efficiently Filtering Table Constraints with Reversible Sparse Bit-Sets}

\author{
Jordan Demeulenaere$^1$, Renaud Hartert$^1$, Christophe Lecoutre$^2$, \\ Guillaume Perez$^3$, Laurent Perron$^4$, Jean-Charles R\'egin$^3$, Pierre Schaus$^1$
}

\institute{$^1$UCLouvain, Belgium, $^2$CRIL, University of Artois $^3$University of Nice, France, $^4$Google, France, }

\date{}

\begin{document}

\thispagestyle{empty}

\maketitle
\vspace{-0,5cm}

\begin{abstract}

In this paper, we describe Compact-Table (CT), a bitwise algorithm to enforce Generalized Arc Consistency (GAC) on table constraints. 
Although this algorithm is the default propagator for table constraints in or-tools and OscaR, two publicly available CP solvers, it has never been described so far. 
Importantly, CT has been recently improved further with the introduction of residues, resetting operations 
and a data-structure called reversible sparse bit-set, used to maintain tables of supports     
(following the idea of tabular reduction): 
tuples are invalidated incrementally on value removals by means of bit-set operations.
The experimentation that we have conducted with OscaR shows that CT outperforms state-of-the-art algorithms STR2, STR3, GAC4R, MDD4R and AC5-TC on standard benchmarks. 
\end{abstract}


\section{Introduction}

Table constraints, also called extension(al) constraints, explicitly express the allowed combinations of values for the variables they involve as sequences of tuples, which are called tables. 
Table constraints can theoretically encode any kind of constraints and are amongst the  most useful ones in Constraint Programming (CP). 
Indeed, they are often required when modeling combinatorial problems in many application fields.
The design of filtering algorithms for such constraints has generated a lot of research effort, see \cite{BR_GAC7,LR_fast,LS_generalized,GJMN_data,U_partition,L_str2,LLY_str3,MVD_optimal,PR_GAC4}.

Over the last decade, many developments have thus been achieved for enforcing the well-known property called Generalized Arc Consistency (GAC) on binary and/or non-binary extensionally defined constraints. 
Among successful techniques, we find:
\begin{itemize}
\item bitwise operations that allow performing parallel operations on bit vectors. Already exploited during the 70's \cite{U_algorithm,G_relational}, they have been applied more recently to the enforcement of arc consistency on binary constraints \cite{B_wordwise,LV_enforcing}.
\item residual supports (residues) that store the last found supports of each value. Initially introduced for ensuring optimal complexity \cite{BRYZ_optimal}, they have been shown efficient in practice \cite{LBH_exploiting,LZBF_arc,LH_study} when used as simple sentinels. 
\item tabular reduction, which is a technique that dynamically maintains the tables of supports. Based on the structure of sparse sets \cite{BT_efficient,LSSL_sparse}, variants of Simple Tabular Reduction (STR) have been proved to be quite competitive \cite{U_partition,L_str2,LLY_str3}. 
\item resetting operations that saves substantial computing efforts in some particular situations. They have been successfully applied to the algorithm GAC4 \cite{PR_GAC4}.
\end{itemize}

In this paper, we introduce a very efficient GAC algorithm for table constraints that combines the use of bitwise operations, residual supports, tabular reduction, and resetting operations.
It is called Compact-Table (CT), and originates from or-tools, the Google solver that won the latest MiniZinc Challenges.
It is important to note that or-tools does not implement many global constraints, but heavily relies on table constraints instead, with CT as embedding propagator. 
Through the years, CT has reached a good level of maturity because it has been continuously improved and extended with many cutting edge ideas such as those introduced earlier. 
Unfortunately, the core algorithm of CT has not been described in the literature so far\footnote{Note that some parts of this paper were published in a Master Thesis report \cite{D_CT}.} and is thus seldom used as a reference for practical comparisons.
The first version of CT implemented in or-tools, with a bit-set representation of tables, dates back to 2012, whereas the version of CT presented in this paper is exactly the last one implemented in OscaR \cite{oscar}.    


\paragraph{Outline.}
After presenting related works in Section \ref{sec:relatedwork}, we introduce some technical background in Section \ref{sec:definitions}. 
Then, we recall in Section \ref{sec:trail} usual state restoration mechanisms implemented in CP solvers, and describe reversible sparse bit-sets in Section \ref{sec:rsbs}.
In Section \ref{sec:ct}, we describe our algorithm CT.
Before concluding, we present in Section \ref{sec:xp} the results of an experimentation we have conducted with CT and its contenders on a large variety of benchmarks. 

\section{Related Work}\label{sec:relatedwork}

Propagators for table constraints are filtering procedures used to enforce GAC.
Given the importance of table constraints, it is not surprising that much research has been carried out in order to find efficient propagators.
This section briefly describes the most efficient ones. 

\paragraph{Generic Algorithms.} On the one hand, GAC3 is a classical general-purpose GAC algorithm \cite{M_AC3} for non-binary constraints. 
Each call to this algorithm for a constraint requires testing if each value is still supported by a valid tuple accepted by the constraint. 
Several improvements to fasten the search for a support gave birth to variants such as GAC2001 \cite{BRYZ_optimal} and GAC3$^{rm}$ \cite{LH_study}.
Unfortunately, the worst-case time complexity of all these algorithms grows exponentially with the arity of the constraints.
On the other hand, GAC4 \cite{MM_good} is a value-based algorithm, meaning here that for each value, it maintains a set of valid tuples supporting it.
Each time a value is removed, all supporting tuples are removed from the associated sets, which allows us to identify values without any more supports. 
GAC4R is a recent improvement of GAC4 \cite{PR_GAC4}, which recomputes the sets of supporting tuples from scratch when it appears to be less costly than updating them based on the removed values.

\paragraph{AC5 Instantiations.}
In \cite{MVD_optimal}, Mairy et al. introduce several instantiations of the generic AC5 algorithm for table constraints, the best of them being AC5-TCOptSparse. 
This algorithm shares some similarities with GAC4 since it precomputes lists of supporting tuples 
which allows us to retrieve efficiently new supports by iterating over these lists. 
Note that a reversible integer is used to indicate the current position of a support in each list. 
This algorithm is implemented in Comet, and has been shown to be efficient on ternary and quaternary constraints. 

\paragraph{Simple Tabular Reduction.}  STR1 \cite{U_partition} and STR2 \cite{L_str2} are coarse-grained algorithms that globally enforce GAC by traversing the constraint tables while dynamically maintaining them: each call to the algorithm for a constraint removes the invalid tuples from its table. 
The improvements brought in STR2 avoid unnecessary operations by considering only relevant subsets of variables when checking the validity of a tuple, and collecting supported values.
Contrary to its predecessors, STR3 \cite{LLY_str3} is a fine-grained (or value-based) algorithm. 
For each value, it initially computes a static array of tuples supporting it, 
and keeps a reversible integer $curr$ that indicates the position of the last valid tuple in the array. 
STR3 also maintains the set of valid tuples. 
STR3 is shown to be complementary to STR2, being more efficient when the tables are not reduced drastically during search. 

\paragraph{Compressed Representations.} Other algorithms gamble on the compression of tables to reduce the time needed to ensure GAC. 
The most promising data structure allowing a more compact representation is the Multi-valued Decision Diagram (MDD) \cite{SKMB_algorithms}.
 Two notable algorithms using MDDs as main data structure are \texttt{mddc} \cite{CY_mdd} and MDD4R \cite{PR_GAC4}. 
The former does not modify the decision diagram and performs a depth-first search of the MDD during propagation to detect which parts of the MDD are consistent or not. MDD4R dynamically maintains the MDD by deleting nodes and edges that do not belong to a solution. 
Each value is matched with its corresponding edges in the MDD, so, when a value has none of its edges present in the MDD, it can be removed. 


\section{Technical Background}\label{sec:definitions}

A {\em constraint network} (CN) $N$ is composed of a set of $n$ variables and a set of $e$ constraints. 
Each {\em variable} $x$ has an associated domain, denoted by $dom(x)$, that contains the finite set of values that can be assigned to it.
Each {\em constraint} $c$ involves an ordered set of variables, called the {\em scope} of $c$ and denoted by $scp(c)$,
and is semantically defined by a {\em relation}, denoted by $rel(c)$, which contains the set of tuples allowed for the variables involved in $c$.
The arity of a constraint $c$ is $|scp(c)|$, i.e., the number of variables involved in $c$.
A (positive) {\em table constraint} $c$ is a constraint such that $rel(c)$ is defined explicitly by listing the tuples that are allowed by $c$.

\begin{example}
The constraint $x \neq y$ with $x \in \{1, 2, 3\}$ and $y \in \{1, 2\}$ can be alternatively defined by the table constraint $c$ such that $scp(c)=\{x,y\}$ and $rel(c)=\{(1, 2), (2, 1), (3, 1), (3,2)\}$. We also write: 
$$ \langle x, y \rangle \in T \quad \text{with} \quad T = \langle (1, 2), (2, 1), (3, 1), (3,2) \rangle$$
\end{example}

Let $\tau = (a_1,a_2,\dots,a_r)$ be a tuple of values associated with an ordered set of variables $X = \{x_1,x_2,\dots,x_r\}$.
The ith value of $\tau$ is denoted by $\tau[i]$ or $\tau[x_i]$.
The tuple $\tau$ is valid iff $\forall i \in 1..r, \tau[i] \in dom(x_i)$.
An $r$-tuple $\tau$ is a {\em support} on the $r$-ary constraint $c$ iff $\tau$ is a valid tuple that is allowed by $c$.
If $\tau$ is a support on a constraint $c$ involving a variable $x$ and such that $\tau[x]=a$, we say that $\tau$ is a {\em support for} $(x,a)$ on $c$.
Generalized Arc Consistency (GAC) is a well-known domain-filtering consistency defined as follows:

\begin{definition} 
A constraint $c$ is \emph{generalized arc consistent} (GAC) iff $\forall x \in scp(c), \forall a \in dom(x)$, there exists at least one support for $(x,a)$ on $c$. 
A CN $N$ is GAC iff every constraint of $N$ is GAC.
\end{definition}

Enforcing GAC is the task of removing from domains all values that have no support on a constraint.  Many algorithms have been devised
for establishing GAC according to the nature of the constraints.
For table constraints, STR \cite{U_partition} is such an algorithm: it removes invalid tuples during search of supports using a sparse set data structure which separates valid tuples from invalid ones. 
This method of seeking supports improves search time by avoiding redundant tests on invalid tuples that have already been detected as invalid during previous GAC enforcements. 
STR2 \cite{L_str2}, an optimization of STR, limits some basic operations concerning the validity of tuples and the identification of supports, through the introduction of two important sets called $S^{sup}$ and $S^{val}$ (described later).

\section{Reversible Objects and Implementation Details}\label{sec:trail}

\paragraph{Trail and Timestamping.}

The issue of storing related states of the solving process is essential in CP.
In many solvers\footnote{One notable exception is Gecode, a copy-based solver.}, a general mechanism is used for doing and undoing (on backtrack) the current state.
This mechanism is called a trail and it was first introduced in \cite{F_nondeterministic} for implementing non-deterministic search.
A trail is a stack of pairs $(location,value)$ where $location$ stands for any piece of memory (e.g., a variable), which can be restored when backtracking.
Typically, at each search node encountered during the solving process, the constraint propagation algorithm is executed. 
A same filtering procedure (propagator) can thus be executed several times at a given node.
Consequently, if one is interested in storing some information concerning a filtering procedure, the value of a same memory location can be changed several times. 
However, stamping that is part of the "folklore" of programming \cite{K_art} can be used to avoid storing a same memory location on the trail more than once per search node.
The idea behind timestamping is that only the final state of a memory location is relevant for its restoration on backtrack.
The trail contains a general time counter that is incremented at each search node, and a timestamp is attached to each memory location indicating the time at which its last storage on the trail happened.
If a memory location changes and its timestamp matches the current time of the trail then there is no need to store it again. 
CP solvers generally expose some "reversible" objects to the users using this trail+timestamping mechanism.
The most basic one is the reversible version of primitive types such as int or long values.
In the following, we denote by \texttt{rint} and \texttt{rlong} the reversible versions of \texttt{int} and \texttt{long} primitive types.

\paragraph{Reversible Sparse Sets.}

Reversible primitive types can be used to implement more complex data structures such as reversible sets.
It was shown in \cite{LSSL_sparse} how to implement a reversible set using a single \texttt{rint} that represents the current size (limit) of the set.
In this structure, which is called reversible sparse set, an array of size $n$ is used to store the permutation from $0$ to $n-1$. 
All values in this permutation array at indices smaller than or equal to a variable $limit$ are considered as part of the set, while the others are considered as removed.
When iterating on current values of the set (with decreasing indices from $limit$ to 0), the value at the current index can be removed in $O(1)$ by just swapping it with 
the value stored at $limit$ and decrementing $limit$.
Making a sparse set reversible just requires managing a single \texttt{rint} for $limit$.
On backtrack, when the limit is restored, all concerned removed values are restored in $O(1)$.

\paragraph{Domains and Deltas.}

In OscaR \cite{oscar}, the implementation of domains relies on reversible sparse sets.
One advantage of implementing domains with this structure is that one can easily retrieve the set of values removed from a domain between any two calls to a given filtering procedure.
All we need to store in the filtering procedure is the last size of the domain.
The delta set (set of values removed between the two calls) is composed of all the values located between the current size and the last recorded size. 
More details on this cheap mechanism to retrieve the delta sets can be found in \cite{LSSL_sparse}.

\section{Reversible Sparse Bit-Sets}\label{sec:rsbs}

This section describes the class \texttt{RSparseBitSet} that is the main data structure for our algorithm to maintain the supports.
In what follows, when we refer to an array $t$, t[0] denotes the first element (indexing starts at 0) and t.length the number of its cells (size).

\SetKwBlock{isEmpty}{Method \m{isEmpty}(): Boolean}{}
\SetKwBlock{clearMask}{Method \m{clearMask}()}{}
\SetKwBlock{reverseMask}{Method \m{reverseMask}()}{}
\SetKwBlock{addToMask}{Method \m{addToMask}(m: array of long)}{}
\SetKwBlock{intersectWithMask}{Method \m{intersectWithMask}()}{}
\SetKwBlock{intersectIndex}{Method \m{intersectIndex}(m: array of long): int}{}

\begin{algorithm}[h!]
  $\f{words}$: array of rlong \tcp*[r]{words.length = p}
  $\f{index}$: array of int  \tcp*[r]{index.length = p}
  $\f{limit}$: rint  \;
  $\f{mask}$: array of long  \tcp*[r]{mask.length = p}
  \BlankLine 
  \isEmpty {
    \Return{$\f{limit} = -1$} \; 
  }
  \BlankLine
  \clearMask { 
    \ForEach{$i$ {\bf from} $0$ {\bf to} $\f{limit}$} {
      $\f{offset} \gets \f{index}[i]$ \;
      $\f{mask}[\f{offset}] \gets \Zero$ \;
    } 
  }
  \BlankLine
  \reverseMask {
    \ForEach{$i$ {\bf from} $0$ {\bf to} $\f{limit}$} {
      $\f{offset} \gets \f{index}[i]$ \;
      $\f{mask}[\f{offset}] \gets$ \textasciitilde $\f{mask}[\f{offset}]$  \tcp*[r]{bitwise NOT}
    }
  }
  \BlankLine
  \addToMask {
    \ForEach{$i$ {\bf from} $0$ {\bf to} $\f{limit}$} {
      $\f{offset} \gets \f{index}[i]$ \;
      $\f{mask}[\f{offset}] \gets \f{mask}[\f{offset}]$ | $m[\f{offset}]$ \tcp*[r]{bitwise OR}
    }
  }
  \BlankLine
  \intersectWithMask {
    \ForEach{$i$ {\bf from} $\f{limit}$ {\bf downto} $0$} {
      $\f{offset} \gets \f{index}[i]$ \;
      $w \gets \f{words}[\f{offset}]$ \& $\f{mask}[\f{offset}]$  \tcp*[r]{bitwise AND}
      \If{$w \neq \f{words}[\f{offset}]$} {
        $\f{words}[\f{offset}] \gets w$ \;    
        \If{$w = \Zero$} {
          $\f{index}[i] \gets \f{index}[\f{limit}]$ \;
          $\f{index}[\f{limit}] \gets \f{offset}$ \;
          $\f{limit} \gets \f{limit} -1$ \;
        } 
      }
    }
  } 
  \BlankLine
  \intersectIndex {
    \tcc*[l]{Post: returns the index of a word where the bit-set intersects with m, -1 otherwise}
    \ForEach{$i$ {\bf from} $0$ {\bf to} $\f{limit}$} {
      $\f{offset} \gets \f{index}[i]$ \;
      \If {$\f{words}[\f{offset}]$ \textsc{\&} $m[\f{offset}] \neq \Zero$} {
        \Return{$\f{offset}$} \;
      }
    }
    \Return{$-1$} \;
}
  \caption{Class RSparseBitSet\label{alg:class1}} 
\end{algorithm}

The class \texttt{RSparseBitSet}, which encapsulates four fields and 6 methods, is given in Algorithm \ref{alg:class1}.
One important field is $\f{words}$, an array of $p$ 64-bit words (actually, reversible long integers), which defines the current value of the bit-set: the ith bit of the jth word is 1 iff the $(j-1) \times 64 + i$th element of the (initial) set is present.  
Initially, all words in this array have all their bits at 1, except for the last word that may involve a suffix of bits at 0. 
For example, if we want to handle a set initially containing 82 elements, then we build an array with $p= \lceil 82/64 \rceil = 2$ words that initially looks like:

\begin{small}\begin{quote}
$\f{words}$: 11111111111111111111111111111111  11111111111111111100000000000000
\end{quote}\end{small}

Because, in our context, only non-zero words (words having at least one bit set to 1) are relevant when processing operations on the bit-set, we rely on the sparse-set technique by managing in an array $\f{index}$ the indices of all words: the indices of all non-zero words are in $\f{index}$ at positions less than or equal to the value of a variable $\f{limit}$, and the indices of all zero-words are in $\f{index}$ at positions strictly greater than $\f{limit}$.  
For our example, we initially have:

\begin{small}\begin{quote}
$\f{words}$: 11111111111111111111111111111111  11111111111111111100000000000000\\
$\f{index}$: 0 1\\
$\f{limit}$ : 1
\end{quote}\end{small}

If we suppose now that the 66 first elements of our set above are removed, we obtain:
\begin{small}\begin{quote}
$\f{words}$:  00000000000000000000000000000000 00111111111111111100000000000000\\
$\f{index}$: 1 0\\
$\f{limit}$: 0
\end{quote}\end{small}

The class invariant describing the state of a reversible sparse bit-set is the following:
\begin{itemize}
\item $\f{index}$ is a permutation of $[0,\ldots,p-1]$, and
\item $\f{words}[\f{index}[i]] \neq \Zero \Leftrightarrow i \leq \f{limit}$, $\forall i \in 0..p-1$
\end{itemize}

Note that the reversible nature of our object comes from 1) an array of reversible long (denoted \texttt{rlong}) (instead of simple longs) to store the bit words, and 2) the reversible prefix size of non-zero words by using a reversible int (\texttt{rint}). 

A \texttt{RSparseBitSet} also contains a kind of local temporary array, called $\f{mask}$. Is is used to collect elements with Method \m{addToMask}(), and can be cleared and reversed too.
A \texttt{RSparseBitSet} can only be modified by means of the method \m{intersectWithMask}() which is an operation used to intersect with the elements collected in $\f{mask}$.
An illustration of the usage of these methods is given in next example.

\begin{figure}
\centering
\begin{tabular}{r|c|c|c|c|c|c|c|c|}
\cline{2-9}
$\f{words}$ & 1 & 0 & 1 & 0 & 1 & 1 & 1 & 1 \\ 
\cline{2-9}
\m{addToMask} & 1 & 1 & 1 & 0 & 1 & 0 & 0 & 0 \\ 
\cline{2-9}
\m{addToMask} & 0 & 0 & 0 & 1 & 0 & 0 & 0 & 1 \\ 
\cline{2-9}
$\f{mask}$      & 1 & 1 & 1 & 1 & 1 & 0 & 0 & 1 \\
\cline{2-9}
\m{intersectWithMask} & 1 & 0 & 1 & 0 & 1 & 0 & 0 & 1 \\ 
\cline{2-9}
\end{tabular} 
\caption{\texttt{RSparseBitSet} example}
\label{fig:rbitset}
\end{figure}

\begin{example}
Figure \ref{fig:rbitset} illustrates the use of Methods \m{addToMask}() and \m{intersectWithMask}().
We assume that the current state of the bit-set is given by the value of $\f{words}$, and that \m{clearMask}() has been called such that $\f{mask}$ is initially empty.
Then two bit-sets are collected in $\f{mask}$ by calling \m{addToMask}().
The value of $\f{mask}$ is represented after these two operations. 
Finally \m{intersectWithMask}() is executed and the new value of the bit-set $\f{words}$ is given at the last row of Figure \ref{fig:rbitset}. 
\end{example}

We now describe the implementation of the methods in \texttt{RSparseBitSet}.
Method \m{isEmpty}() simply checks if the number of non-zero words is different from zero (if the limit is set to -1, it means that all words are non-zero).
Method \m{clearMask}() sets to 0 all words of $\f{mask}$ corresponding to non-zero words of $\f{words}$, whereas Method \m{reverseMask}() reverses all words of $\f{mask}$.
Method \m{addToMask}() applies a word by word logical bit-wise \emph{or} operation. Once again, notice that this operation is only applied to words of $\f{mask}$ corresponding to non-zero words of $\f{words}$. 
Method \m{intersectMask}() considers each non-zero word of $\f{words}$ in turn and replaces it by its intersection with the corresponding word of $\f{mask}$.
In case the resulting new word is zero, it is swapped with the last non-zero word and the value of $\f{limit}$ is decremented. 
Finally, Method \m{intersectIndex}() checks if a given bit-set (array of longs) intersects with the current bit-set: it returns the index of the first word where an intersection can be proved, -1 otherwise.


\section{Compact-Table (CT) Algorithm}\label{sec:ct}

As STR2 and STR3, Compact-Table (CT) is a GAC algorithm that dynamically maintains the set of valid supports regarding the current domain of each variable. 
The main difference is that CT is based on an object \texttt{RSparseBitSet}.
In this set, each tuple is indexed by the order it appears in the initial table.
Invalid tuples are removed during the initialization as well as values that are not supported by any tuple.
The class \texttt{ConstraintCT}, Algorithm \ref{alg:class2}, allows us to implement any positive table constraint $c$ while running the CT algorithm to enforce GAC.

\subsection{Fields}

As fields of Class \texttt{ConstraintCT}, we first find $\f{scp}$ for representing the scope of $c$ and $\f{currTable}$ for representing the current table of $c$ by means of a reversible sparse bit-set.
If $\langle \tau_0, \tau_1, \dots, \tau_{p-1} \rangle$ is the initial table of $c$, then $\f{currTable}$ is a \texttt{RSparseBitSet} object (of initial size $p$) such that the value $i$ is contained (is set to 1) in the bit-set if and only if the $i$th tuple is valid:
$$i \in \f{currTable} \Leftrightarrow \forall x \in scp(c), \tau_i[x] \in dom(x)$$

We also have three fields $\f{S^{val}}$, $\f{S^{val}}$ and $\f{lastSizes}$ in the spirit of STR2. 
Indeed, as in \cite{L_str2}, we introduce two sets of variables, called $\f{S^{val}}$ and $\f{S^{sup}}$.
The set $\f{S^{val}}$ contains uninstantiated variables (and possibly, the last assigned variable) whose domains have been reduced since the previous invocation of the filtering algorithm on $c$.
To set up $\f{S^{val}}$, we need to record the domain size of each modified variable $x$ right after the execution of CT on $c$: this value is recorded in $\f{lastSizes}[x]$. 
The set $\f{S^{sup}}$ contains uninstantiated variables (from the scope of the constraint $c$) whose domains contain each at least one value for which a support must be found.   
These two sets allow us to restrict loops on variables to relevant ones.

\SetKwBlock{updateTable}{Method \m{updateTable}()}{}
\SetKwBlock{filterDomains}{Method \m{filterDomains}()}{}
\SetKwBlock{enforceGAC}{Method \m{enforceGAC}()}{}

\begin{algorithm}[h!]
  $\f{scp}$: array of variables  \tcp*[r]{Scope}
  $\f{currTable}$: RSparseBitSet  \tcp*[r]{Current table}
  $\f{S^{val}}$, $\f{S^{sup}}$ \tcp*[r]{Temporary sets of variables} 
  $\f{lastSizes}$ \tcp*[r]{$\f{lastSizes}[x]$ is the last size of the domain of $x$} 
  $\f{supports}$ \tcp*[r]{$\f{supports}[x,a]$ is the bit-set of supports for $(x,a)$} 
  $\f{residues}$ \tcp*[r]{$\f{residues}[x,a]$ is the last found support for $(x,a)$}
  \BlankLine 
  \updateTable {
    \ForEach{variable $x \in \f{S^{val}}$} {  \label{line:xiter1}
    	  $\f{currTable}.\m{clearMask}$() \;
      \uIf(\tcp*[f]{Incremental update}){$|\Delta_x| < |dom(x)|$} { \label{line:delta}  
        \ForEach{value $a \in \Delta_x$} {
          $\f{currTable}.\m{addToMask}(\f{supports}[x,a])$ \;
        }
        $\f{currTable}.\m{reverseMask}()$ \;
        
      } 
      \Else(\tcp*[f]{Reset-based update}) { \label{line:reset}  
        \ForEach{value $a \in dom(x)$} {
          $\f{currTable}.\m{addToMask}(\f{supports}[x,a])$ \;
        }
      }
      $\f{currTable}.\m{intersectWithMask}()$ \;
      \If{$\f{currTable}.\m{isEmpty}()$} {
      	{\bf break} \;
      }
    }
  }
  \BlankLine
  \filterDomains { 
    \ForEach{variable $x \in \f{S^{sup}}$} { \label{line:xiter2}
      \ForEach{value $a \in dom(x)$} {
        $\f{index} \gets \f{residues}[x,a]$ \;
        \If{$\f{currTable}.\f{words}[\f{index}]$ \textsc{\&} $\f{supports}(x,a)[\f{index}] = \Zero$} {
          $\f{index} \gets \f{currTable}.\m{intersectIndex}(\f{supports}[x,a])$ \;
          \uIf{$\f{index} \neq -1$} {
            $\f{residues}[x,a] \gets \f{index}$ \;
          } \Else {
            $dom(x) \gets dom(x) \setminus \{a\}$ \;
          }
        }
      }
      $\f{lastSize}[x] \gets |dom(x)|$ \;
    }
  } \BlankLine
  \enforceGAC {
    $\f{S^{val}}  \gets \{x \in \f{scp} : |dom(x)| \neq \f{lastSize}[x]\}$ \; \label{line:sval}
    \ForEach{variable $x \in \f{S^{val}}$} {
      $\f{lastSize}[x] \gets |dom(x)|$\;
    }
    $\f{S^{sup}} \gets \{x \in \f{scp} : |dom(x)|>1\}$ \;
    \m{updateTable}() \; 
    \If{$\f{currTable}.\m{isEmpty}()$} {
      \Return{Backtrack} \;
    }
    \m{filterDomains}() \;
  }
  \caption{Class ConstraintCT\label{alg:class2}} 
\end{algorithm}

We also have a field $\f{supports}$ containing static data.
During the set up of the table constraint $c$, CT also computes a static array of words $\f{supports}[x,a]$, seen as a bit-set, for each variable-value pair $(x,a)$ where $x \in scp(c) \land a \in dom(x)$: the bit at position $i$ in the bit-set is 1 if and only if the tuple $\tau_i$ in the initial table of $c$ is a support for $(x, a)$. 

\input{figures/initmasks.tex}

\begin{example}\label{example:initialization}
Figure \ref{fig:initmasks} shows an illustration of the content of those bit-sets after the initialization of the following table constraint $\langle x,y,z \rangle \in T$, with:
\begin{itemize}
\item $dom(x)=\{ a, b \}$, $dom(y)=\{ a, b, d \}$, $dom(z)=\{ a, b, c \}$
\item $T = \langle (a, a, a), (a, a, b), (a, b, c), (b, a, a), (a, c, b), (a, b, b), (b, a, b), (b, b, a), (b, b, b) \rangle$
\end{itemize}
The tuple $(a, c, b)$ is initially invalid because $c \notin dom(y)$, and thus will not be indexed. Value $d$ will be removed from $dom(y)$ given that it is not supported by any tuple.
\end{example}

Finally, we have an array $\f{residues}$ such that for each variable-value pair $(x,a)$, $\f{residues}[x,a]$ denotes the index of the word where a support was found for $(x,a)$ the last time one was sought for.

\subsection{Methods \label{section:propagation}}

The main method in \texttt{ConstraintCT} is \m{enforceGAC}().
After the initialization of the sets $\f{S^{val}}$ and $\f{S^{sup}}$, CT updates $\f{currTable}$ to filter out (indices of) tuples that are no more supports, and then considers each variable-value pair to check whether these values still have a support. 

\subsubsection{Updating the Current Table}

For each variable $x \in \f{S^{val}}$, i.e., each variable $x$ whose domain has changed since the last time the filtering algorithm was called, \m{updateTable}() performs some operations.
This method assumes an access to the set of values $\Delta_x$ removed from $dom(x)$.
There are two ways of updating $\f{currTable}$, either incrementally or from scratch after resetting.
Note that the idea of using resets has been proposed in \cite{PR_GAC4} and successfully applied to GAC4 and MDD4, with the practical interest of saving computational effort in some precise contexts.
This is the strategy implemented in \m{updateTable}(), by considering a reset-based computation when the size of the domain is smaller than the number of deleted values.

In case of an incremental update (line \ref{line:delta}), the union of the tuples to be removed is collected by calling \m{addToMask}() for each bit-set (of supports) corresponding to removed values, whereas in case of a reset-based update (line \ref{line:reset}), we perform the union of the tuples to be kept. To get a mask ready to apply, we just need to reverse it when it has been built from removed values.
Finally, the (indexes of) tuples of $\f{currTable}$ not contained in the mask, built from $x$, are directly removed by means of \m{intersectWithMask}().
When there is no more tuple in the current table, a failure is detected, and \m{updateTable}() is stopped by means of a loop break.

\subsubsection{Filtering of Domains}

Values are removed from the domain of some variables during the search of a solution, which can lead to inconsistent values in the domain of other variables. 
As $\f{currTable}$ is a reversible and dynamically maintained structure, the value of some bits changes from 1 to 0 when tuples become invalid (or from 0 to 1 when the search backtracks). 
On the contrary, the $\f{supports}$ bit-sets are only computed at the creation of the constraint and are not maintained during search. 
It follows from the definition of those bit-sets that $(x,a)$ has a valid support if and only if
\begin{equation}\label{eq:mask_condition}
\left( \f{currTable} \cap \f{supports}[x, a] \right) \neq \emptyset
\end{equation}

Therefore, each time a tuple becomes invalid, the constraint must check this condition for every variable value pair $(x,a)$ such that $a \in dom(x)$, and remove $a$ from $dom(x)$ if the condition is not satisfied any more. 
This operation is efficiently implemented in \m{filterDomains}() with the help of residues and the method \m{intersectIndex}().

\begin{figure}
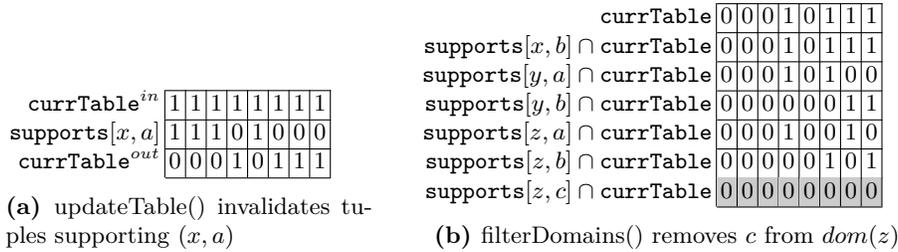

\begin{center} 
\begin{subfigure}[b]{0.4\textwidth}
\begin{tabular}{r|c|c|c|c|c|c|c|c|}
\cline{2-9}
$\f{currTable}^{in}$ & 1 & 1 & 1 & 1 & 1 & 1 & 1 & 1 \\ 
\cline{2-9}
$\f{supports}[x, a]$ & 1 & 1 & 1 & 0 & 1 & 0 & 0 & 0 \\ 
\cline{2-9}
$\f{currTable}^{out}$ & 0 & 0 & 0 & 1 & 0 & 1 & 1 & 1 \\ 
\cline{2-9}
\end{tabular} 
\caption{\m{updateTable}() invalidates tuples supporting $(x,a)$\label{subfig:update_naif}}
\end{subfigure}
\hspace{0.4cm}
\begin{subfigure}[b]{0.55\textwidth}
\captionsetup{justification=centering}
\begin{tabular}{r|c|c|c|c|c|c|c|c|}
\cline{2-9}
$\f{currTable}$ & 0 & 0 & 0 & 1 & 0 & 1 & 1 & 1 \\ 
\cline{2-9}
$\f{supports}[x,b] \cap \f{currTable}$ & 0 & 0 & 0 & 1 & 0 & 1 & 1 & 1 \\  
\cline{2-9}
$\f{supports}[y,a] \cap \f{currTable}$ & 0 & 0 & 0 & 1 & 0 & 1 & 0 & 0 \\ 
\cline{2-9}
$\f{supports}[y,b]\cap \f{currTable}$ & 0 & 0 & 0 & 0 & 0 & 0 & 1 & 1 \\ 
\cline{2-9}
$\f{supports}[z,a] \cap \f{currTable}$ & 0 & 0 & 0 & 1 & 0 & 0 & 1 & 0 \\ 
\cline{2-9}
$\f{supports}[z,b] \cap \f{currTable}$ & 0 & 0 & 0 & 0 & 0 & 1 & 0 & 1 \\ 
\cline{2-9}
$\f{supports}[z,c] \cap \f{currTable}$ & \cellcolor{gray!40}{0} & \cellcolor{gray!40}{0} & \cellcolor{gray!40}{0} & \cellcolor{gray!40}{0} & \cellcolor{gray!40}{0} & \cellcolor{gray!40}{0} & \cellcolor{gray!40}{0} & \cellcolor{gray!40}{0} \\ 
\cline{2-9}
\end{tabular} 
\caption{\m{filterDomains}() removes $c$ from $dom(z)$\label{subfig:propagateZc}}
\end{subfigure}
\caption{Illustration of \m{enforceGAC}() after the removal of $a$ from $dom(x)$.}
\end{center}
\end{figure}

\begin{example}
The same set of tuples as in Example \ref{example:initialization} is considered.
Suppose now that $a$ was removed from $dom(x)$ (by another constraint) after the initialization. 
Given that the domain of $x$ is reduced, when \m{updateTable}() is called by \m{enforceGAC}(), all tuples supporting $a$ (because $\Delta_x = \{a\}$) will be invalidated. 
Figure \ref{subfig:update_naif} illustrates the intermediary bit-sets used to compute the new value $\f{currTable}^{out}$ from $\f{currTable}^{in}$ and $\f{supports}[x,a]$. 
Then \m{filterDomains}() computes for each variable-value pair $(x_i,a_i)$ (with $x_i \in \f{S^{sup}}$ and $a_i \in dom(x)$) the intersection of its associated set of supports with $\f{currTable}$ as shown in Figure \ref{subfig:propagateZc}. 
Given that the intersection for $\f{supports}[z,c]$ and $\f{currTable}$ is empty, $c$ is removed from $dom(z)$.
\end{example}

\subsection{Improvements}
The algorithm in Section \ref{section:propagation} can be improved to avoid unnecessary computations in some cases.

\paragraph{Filtering out bounded variables. \label{subsection:bounded}}
The initialization of $\f{S^{val}}$ at line \ref{line:sval} can be only performed from unbound variables (and the last assigned variable), instead from the whole scope. 
We can maintain them in a reversible sparse set.

\paragraph{Last modified variable. \label{subsection:lastTouched}}
It is not necessary to attempt to filter values out from the domain of a variable $x$ if this was the only modified variable since the last call to \m{enforceGAC}(). 
Indeed, when \m{updateTable}() is executed, the new state of $\f{currTable}$ will be computed from $dom(x)$ or $\Delta_x$ only. 
Because every value of $x$ had a support in $\f{currTable}$ the last time the propagator was called, we can omit filtering $dom(x)$ by initially removing $x$ from  $\f{S^{sup}}$. 

\section{Experiments}\label{sec:xp}

We experimented CT on $1,621$ CSP instances involving (positive) table constraints (15GB of uncompressed files in format XCSP 2.1).
This corresponds to a large variety of instances, taken from 37 series.
For guiding search, we used binary branching with \textit{domain over degree} as variable ordering heuristic and \textit{min value} as value ordering heuristic.
A timeout of $1,000$ seconds was used for each instance.
The tested GAC algorithms are CT, STR2 \cite{L_str2}, STR3 \cite{LLY_str3}, GAC4 \cite{MM_good,PR_GAC4}, GAC4R \cite{PR_GAC4}, MDDR \cite{PR_GAC4} and AC5TCRecomp \cite{MHD_optimal}.
All scripts, codes and benchmarks allowing to reproduce our experiments are available at \url{https://bitbucket.org/pschaus/xp-table}.
The experiments were run on a 32-core machine (1400MHz cpu) with 100GB using Java(TM) SE Runtime Environment (build 1.8.0\_60-b27) with 10GB of memory allocated (-Xmx option).

\paragraph{Performance Profiles.} 
Let $t_{p,\,s}$ represent the time obtained with filtering algorithm $s \in S$ on instance $p \in P$. 
The performance ratio is defined as follows: $r_{p,\,s} = \frac{t_{p,\,s}}{\min\{t_{p,\,s^*} | s^* \in S\}}$, where $s^*$ denotes the fastest algorithm.
 A ratio $r_{p, \,s} = 1$ means that $s$ was the fastest on $p$. 
The performance profile \cite{DM_benchmarking} is a cumulative distribution function of the performance of $s$ compared to other algorithms:
 $$\rho_s(\tau) = \frac{1}{|P|} \times |\{p \in P | r_{p,\,s} \leq \tau\}|$$

\begin{figure}
\includegraphics[width=1\textwidth]{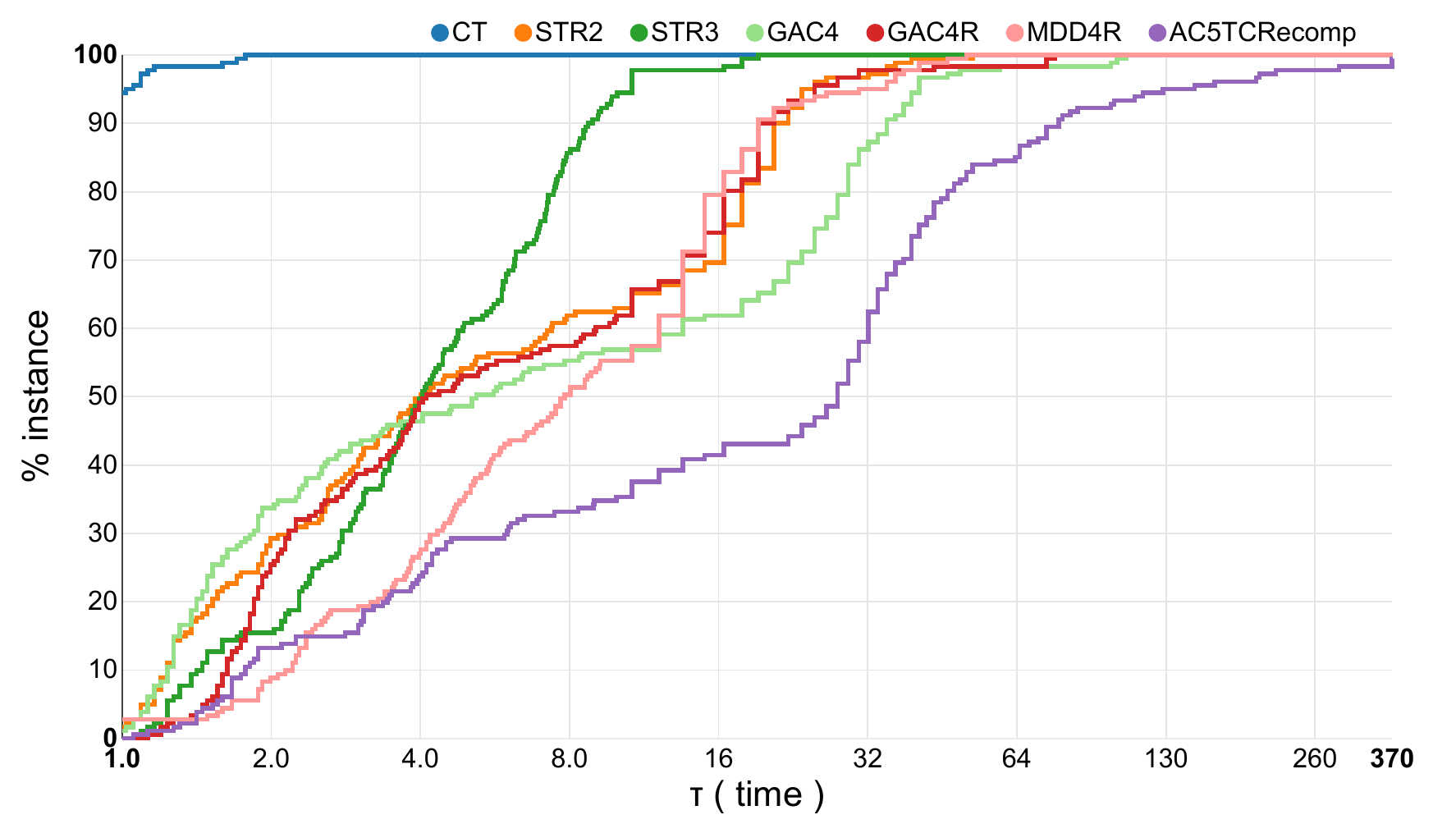}
\caption{Performance Profile}\label{fig:perfprofile}
\end{figure}

Our results are visually aggregated to form a performance profile in Figure \ref{fig:perfprofile} generated by means of the online tool \url{http://sites.uclouvain.be/performance-profile}. 
Note that we filtered out the instances that i) could not be solved within $1,000$ seconds by all algorithms ii) were solved in less than 2 seconds by the slowest algorithm, and iii) required less than 500 backtracks. The final set of instances used to build the profile is composed of 227 instances. An interactive performance profile is also available at \url{https://www.info.ucl.ac.be/~pschaus/assets/publi/performance-profile-ct} to let the interested reader deactivate some family of instances to analyze the results more closely.

\begin{table}[h]
\begin{center}
\begin{tabular}{l S[table-format=5.2] S[table-format=5.2] S[table-format=5.2] S[table-format=5.2] S[table-format=5.2] S[table-format=5.2] S[table-format=5.2] } 
\toprule
Speedup	& {STR2} & {STR3} & {GAC4} & {GAC4R} & {MDD4R} & {AC5-TC} & {Best2} \\ 
\midrule
\rowcolor{vlgray} average	&9.11	&5.07&	15.59&	11.37&	10.38&	50.40	&3.77\\
min	&0.76	&1.09&	0.92	&1.13&	0.13	&1.05&	0.13\\
\rowcolor{vlgray} max	&88.58	&51.04	&173.24	&208.52&	50.84	&1850.82	&15.99\\
std	&10.64	&4.36	&19.67&	18.57&	9.46	&134.13&	2.87\\
\bottomrule
\end{tabular}
\end{center}
\caption{Speedup analysis of CT over the other algorithms. Column 'Best2' corresponds to a virtual second best solver (by considering the minimum time taken by all algorithms except CT).}
\label{table:speedup}
\end{table}

Table \ref{table:speedup} reports the speedup statistics of CT over the other algorithms.
A first observation is that CT is the fastest algorithm on 94.47\% of the instances.
Among all tested algorithms, AC5TCRecomp obtains the worse results.
Then it is not clear which one among STR2, STR3, GAC4 and GAC4R is the second best algorithm.
Based on the average speedup, STR3 seems to be the second best algorithm followed by STR2, MDD4R and GAC4R.
Importantly, one can observe that the speedup of CT over the best of the other algorithms is about 3.77 on average.

\begin{figure}
\includegraphics[width=1\textwidth]{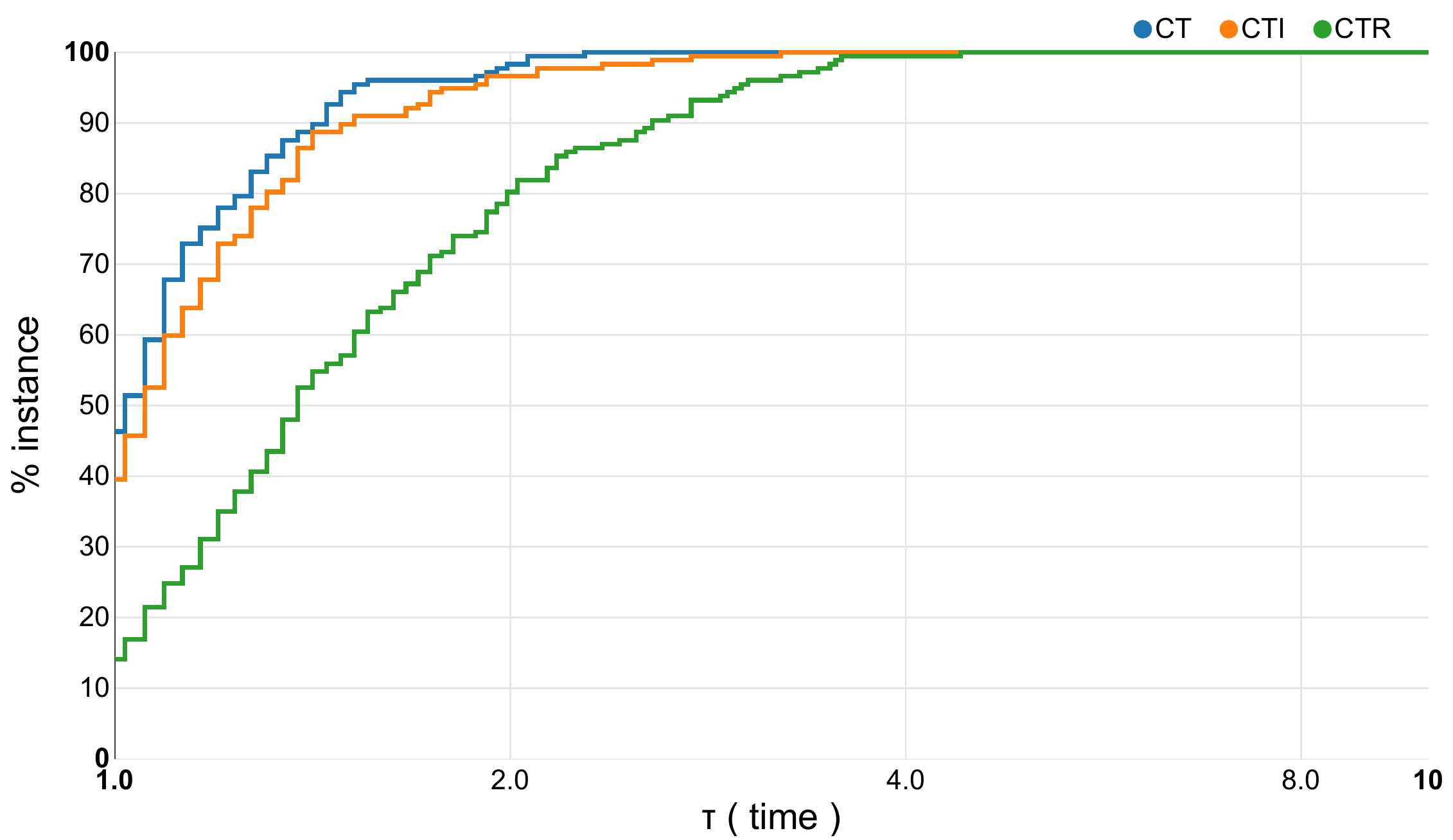}
\caption{Performance Profiles with dynamic (CT), recomputation (CTR) and incremental (CTI) strategies.}
\label{fig:perfprofilect}
\end{figure}

\paragraph{Impact of Resetting Operations.}
In Algorithm \ref{alg:class2}, the choice of being incremental or not, when updating \texttt{currTable}, depends on the size of several sets and is thus dynamic.
We propose to analyze two variants of Algorithm \ref{alg:class2} when this choice is static:
\begin{itemize}
\item Full incremental (CTI): only the body of the 'if' at line \ref{line:delta} is executed (deltas are systematically used).
\item Full re-computation (CTR) : only the body of the 'else' at line \ref{line:reset} is executed (domains are systematically used).
\end{itemize}
The performance profiles with these two variants are given in Figure \ref{fig:perfprofilect}, and the speedup table of the static versions over the dynamic one is given in Table \ref{table:speedupct}.

\begin{table}[h]
\begin{center}
\begin{tabular}{l S[table-format=4.2] S[table-format=4.2] S[table-format=4.2]} 
\toprule
Speedup	& {CTI} & {CTR} & {Best} \\ 
\midrule
\rowcolor{vlgray} average	& 1.09	& 1.46 & 1.02\\
min	& 0.44	 & 0.53	& 0.44 \\
\rowcolor{vlgray} max	& 3.23	&4.39	& 1.96\\
std	&0.38& 0.65& 	0.27\\
\bottomrule
\end{tabular}
\end{center}
\caption{Speedup analysis of the two static variants over CT.} 
\label{table:speedupct}
\end{table}

As can be seen from both the performance profiles and the speedup table, the dynamic version using the resetting operations as introduced in \cite{PR_GAC4} dominates the static ones. 
The average speedup is around 9\% over CTI and 46\% over CTR. 

\paragraph{Contradiction with Previous Results.}
In \cite{MHD_optimal}, AC5TCRecomp was presented as being competitive with STR2. 
When we analyzed the code\footnote{available at \url{http://becool.info.ucl.ac.be}} of STR2 used in \cite{MHD_optimal}, it appeared that STR2 was implemented in Comet using built-in sets (triggering the garbage collection of Comet). 
We thus believe that the results and conclusions in \cite{MHD_optimal} may over-penalize the performance of STR2.
Our results also somehow contradicts the results in \cite{PR_GAC4} where STR3 and STR2 were dominated by MDD4R and GAC4R.
When analyzing the performance of the implementation of STR2 and STR3 used in \cite{PR_GAC4} with or-tools, it appears that it is not as competitive as that in AbsCon (sometimes slower by a factor of 3). 
The results presented in \cite{PR_GAC4} may thus also over-penalize the performances of STR2 and STR3.

One additional contribution of this work is a fined-tuned implementation of the best filtering algorithms for table constraints. 
The implementation of all these algorithms in OscaR was optimized, and checked to be close in performance to the ones by the original authors. 
For CT, STR2 and STR3, a comparison was made with AbsCon, and for CT, MDD4R and GAC4R, a comparison was made with or-tools.
Our implementation required a development effort of 10 man-months in order to obtain an efficient implementation of each algorithm. 
It involved the expertise of several OscaR developers and a deep analysis of the existing implementations in AbsCon and or-tools.
The implementation of all table algorithms used in this paper is open-source and available in OscaR release 3.1.0.

\section{Conclusion}\label{sec:conclusion}

In this paper, we have shown that Compact-Table (CT) is a robust algorithm that clearly dominates state-of-the-art propagators for table constraints.
CT benefits from well-tried techniques: bitwise operations, residual supports, tabular reduction and resetting operations.
We believe that CT can be easily implemented using the reversible sparse bit-set data structure.

\bibliographystyle{plain}
\bibliography{biblio}

\end{document}

%% file: figures/initmasks.tex
\begin{figure}
\centering
\begin{subfigure}[b]{0.4\textwidth}
\centering
\begin{tabular}{|c|c|c|c|}
\hline 
\textbf{T} & $x$ & $y$ & $z$ \\ 
\hline 
0 & a & a & a \\ 
\hline 
1 & a & a & b \\ 
\hline 
2 & a & b & c \\ 
\hline 
3 & b & a & a \\ 
\hline 
\rowcolor{gray!40}
 & a & c & b \\ 
\hline 
4 & a & b & b \\ 
\hline 
5 & b & a & b \\ 
\hline 
6 & b & b & a \\ 
\hline 
7 & b & b & b \\ 
\hline 
\end{tabular} 
\caption{The indexed tuples}
\end{subfigure}
\begin{subfigure}[b]{0.59\textwidth}
\centering
\begin{tabular}{r|c|c|c|c|c|c|c|c|}
\cline{2-9}
$\f{currTable}$ & 1 & 1 & 1 & 1 & 1 & 1 & 1 & 1 \\ 
\cline{2-9} 
$\f{supports}[x,a]$ & 1 & 1 & 1 & 0 & 1 & 0 & 0 & 0 \\ 
\cline{2-9}
$\f{supports}[x,b]$ & 0 & 0 & 0 & 1 & 0 & 1 & 1 & 1 \\ 
\cline{2-9} 
$\f{supports}[y,a]$ & 1 & 1 & 0 & 1 & 0 & 1 & 0 & 0 \\ 
\cline{2-9}
$\f{supports}[y,b]$ & 0 & 0 & 1 & 0 & 1 & 0 & 1 & 1 \\ 
\cline{2-9} 
$\f{supports}[y,d]$ & \cellcolor{gray!40}{0} & \cellcolor{gray!40}{0} & \cellcolor{gray!40}{0} & \cellcolor{gray!40}{0} & \cellcolor{gray!40}{0} & \cellcolor{gray!40}{0} & \cellcolor{gray!40}{0} & \cellcolor{gray!40}{0} \\ 
\cline{2-9} 
$\f{supports}[z,a]$ & 1 & 0 & 0 & 1 & 0 & 0 & 1 & 0 \\ 
\cline{2-9}
$\f{supports}[z,b]$ & 0 & 1 & 0 & 0 & 1 & 1 & 0 & 1 \\ 
\cline{2-9}
$\f{supports}[z,c]$ & 0 & 0 & 1 & 0 & 0 & 0 & 0 & 0 \\ 
\cline{2-9}
\end{tabular} 
\caption{The corresponding bit-sets}
\end{subfigure}
\caption{Illustration of the data structures after the initialization of $\langle x,y,z \rangle \in T$. The tuple $(a, c, b)$ will not be indexed and $d$ will be removed from $dom(y)$.\label{fig:initmasks}}
\end{figure}

%% file: paper.bbl
\begin{thebibliography}{10}

\bibitem{BR_GAC7}
C.~Bessiere and J.-C. R{\'e}gin.
\newblock Arc consistency for general constraint networks: preliminary results.
\newblock In {\em Proceedings of IJCAI'97}, pages 398--404, 1997.

\bibitem{BRYZ_optimal}
C.~Bessiere, J.-C. R\'egin, R.~Yap, and Y.~Zhang.
\newblock An optimal coarse-grained arc consistency algorithm.
\newblock {\em Artificial Intelligence}, 165(2):165--185, 2005.

\bibitem{B_wordwise}
C.~Bliek.
\newblock Wordwise algorithms and improved heuristics for solving hard
  constraint satisfaction problems.
\newblock Technical Report 12-96-R045, ERCIM, 1996.

\bibitem{BT_efficient}
P.~Briggs and L.~Torczon.
\newblock An efficient representation for sparse sets.
\newblock {\em ACM Letters on Programming Languages and Systems},
  2(1-4):59--69, 1993.

\bibitem{CY_mdd}
K.~Cheng and R.~Yap.
\newblock An {MDD}-based generalized arc consistency algorithm for positive and
  negative table constraints and some global constraints.
\newblock {\em Constraints}, 15(2):265--304, 2010.

\bibitem{D_CT}
J.~Demeulenaere.
\newblock Efficient algorithms for table constraints.
\newblock Technical report, Master Thesis, under the supervision of P. Schauss,
  UCLouvain, 2015.

\bibitem{DM_benchmarking}
E.D. Dolan and J.J. Mor{\'e}.
\newblock Benchmarking optimization software with performance profiles.
\newblock {\em Mathematical programming}, 91(2):201--213, 2002.

\bibitem{F_nondeterministic}
R.W. Floyd.
\newblock Nondeterministic algorithms.
\newblock {\em Journal of the ACM}, 14(4):636--644, 1967.

\bibitem{GJMN_data}
I.P. Gent, C.~Jefferson, I.~Miguel, and P.~Nightingale.
\newblock Data structures for generalised arc consistency for extensional
  constraints.
\newblock In {\em Proceedings of AAAI'07}, pages 191--197, 2007.

\bibitem{MVD_optimal}
P.~Van~Hentenryck J.-B.~Mairy and Y.~Deville.
\newblock Optimal and efficient filtering algorithms for table constraints.
\newblock {\em Constraints}, 19(1):77--120, 2014.

\bibitem{K_art}
D.E. Knuth.
\newblock {\em The Art of Computer: Combinatorial Algorithms}, volume~4.
\newblock Addison-Wesley, 2015.

\bibitem{LSSL_sparse}
V.~le~Cl{\'e}ment~de Saint-Marcq, P.~Schaus, C.~Solnon, and C.~Lecoutre.
\newblock Sparse-sets for domain implementation.
\newblock In {\em Proceeding of TRICS'13}, pages 1--10, 2013.

\bibitem{L_str2}
C.~Lecoutre.
\newblock {STR2}: Optimized simple tabular reduction for table constraints.
\newblock {\em Constraints}, 16(4):341--371, 2011.

\bibitem{LBH_exploiting}
C.~Lecoutre, F.~Boussemart, and F.~Hemery.
\newblock Exploiting multidirectionality in coarse-grained arc consistency
  algorithms.
\newblock In {\em Proceedings of CP'03}, pages 480--494, 2003.

\bibitem{LH_study}
C.~Lecoutre and F.~Hemery.
\newblock A study of residual supports in arc consistency.
\newblock In {\em Proceedings of IJCAI'07}, pages 125--130, 2007.

\bibitem{LLY_str3}
C.~Lecoutre, C.~Likitvivatanavong, and R.~Yap.
\newblock {STR3}: A path-optimal filtering algorithm for table constraints.
\newblock {\em Artificial Intelligence}, 220:1--27, 2015.

\bibitem{LS_generalized}
C.~Lecoutre and R.~Szymanek.
\newblock Generalized arc consistency for positive table constraints.
\newblock In {\em Proceedings of CP'06}, pages 284--298, 2006.

\bibitem{LV_enforcing}
C.~Lecoutre and J.~Vion.
\newblock Enforcing arc consistency using bitwise operations.
\newblock {\em Constraint Programming Letters}, 2:21--35, 2008.

\bibitem{LR_fast}
O.~Lhomme and J.-C. R\'egin.
\newblock A fast arc consistency algorithm for n-ary constraints.
\newblock In {\em Proceedings of AAAI'05}, pages 405--410, 2005.

\bibitem{LZBF_arc}
C.~Likitvivatanavong, Y.~Zhang, J.~Bowen, and E.C. Freuder.
\newblock Arc consistency in {MAC}: a new perspective.
\newblock In {\em Proceedings of CPAI'04 workshop held with CP'04}, pages
  93--107, 2004.

\bibitem{M_AC3}
A.K. Mackworth.
\newblock Consistency in networks of relations.
\newblock {\em Artificial Intelligence}, 8(1):99--118, 1977.

\bibitem{MHD_optimal}
J.-B. Mairy, P.~van Hentenryck, and Y.~Deville.
\newblock An optimal filtering algorithm for table constraints.
\newblock In {\em Proceedings of CP'12}, pages 496--511, 2012.

\bibitem{G_relational}
J.J. McGregor.
\newblock Relational consistency algorithms and their application in finding
  subgraph and graph isomorphisms.
\newblock {\em Information Sciences}, 19:229--250, 1979.

\bibitem{MM_good}
R.~Mohr and G.~Masini.
\newblock Good old discrete relaxation.
\newblock In {\em Proceedings of ECAI'88}, pages 651--656, 1988.

\bibitem{oscar}
{OscaR Team}.
\newblock {O}sca{R}: {S}cala in {O}{R}, 2012.
\newblock Available from \url{https://bitbucket.org/oscarlib/oscar}.

\bibitem{PR_GAC4}
G.~Perez and J.-C. R\'egin.
\newblock Improving {GAC-4 for Table and MDD} constraints.
\newblock In {\em Proceedings of CP'14}, pages 606--621, 2014.

\bibitem{SKMB_algorithms}
A.~Srinivasan, T.~Kam, S.~Malik, and R.K. Brayton.
\newblock Algorithms for discrete function manipulation.
\newblock In {\em Proceedings of ICCAD'90}, pages 92--95, 1990.

\bibitem{U_algorithm}
J.R. Ullmann.
\newblock An algorithm for subgraph isomorphism.
\newblock {\em Journal of the ACM}, 23(1):31--42, 1976.

\bibitem{U_partition}
J.R. Ullmann.
\newblock Partition search for non-binary constraint satisfaction.
\newblock {\em Information Science}, 177:3639--3678, 2007.

\end{thebibliography}
